\documentclass[preprint,12pt]{elsarticle}

 \makeatletter
 \def\ps@pprintTitle{%
  \let\@oddhead\@empty
  \let\@evenhead\@empty
  \def\@oddfoot{}%
  \let\@evenfoot\@oddfoot}
 \makeatother

\usepackage{graphicx}
\usepackage{amssymb}

\usepackage{lineno}

\usepackage{booktabs}
\usepackage[load-configurations=version-1]{siunitx} 
\usepackage{longtable}
\usepackage{hhline}
\usepackage{wrapfig}
\usepackage{tabularx}
\usepackage{cancel}
\usepackage{ulem}
\usepackage{natbib}
\usepackage{xcolor}

\usepackage[font=small,labelfont=bf]{caption}

\newcommand{\acks}[1]{\section*{Acknowledgments}#1}

\begin{document}

\begin{frontmatter}


\title{Enriched Annotations for Tumor Attribute Classification from Pathology Reports with Limited Labeled Data}



\author[label5,label1]{Nick Altieri}
\author[label5,label1]{Briton Park}
\author[label2]{Mara Olson}
\author[label3]{John DeNero}
\author[label4]{Anobel Odisho}
\author[label1,label3]{Bin Yu}

\fntext[label5]{Equal Contribution}

\address[label1]{University of California Berkeley, Department of Statistics}
\address[label2]{University of California San Francisco, School of Medicine}
\address[label3]{University of California Berkeley, EECS}
\address[label4]{University of California San Francisco, Department of Urology}

\begin{abstract}

Precision medicine has the potential to revolutionize healthcare, but much of the data for patients is locked away in unstructured free-text, limiting research and delivery of effective personalized treatments. Generating large annotated datasets for information extraction from clinical notes is often challenging and expensive due to the high level of expertise needed for high quality annotations. To enable natural language processing for small dataset sizes, we develop a novel enriched hierarchical annotation scheme and algorithm, Supervised Line Attention (SLA), and apply this algorithm to predicting categorical tumor attributes from kidney and colon cancer pathology reports from the University of California San Francisco (UCSF). Whereas previous work only annotated document level labels, we in addition ask the annotators to enrich the traditional label by asking them to also highlight the relevant line or potentially lines for the final label, which leads to a 20\% increase of annotation time required per document. With the enriched annotations, we develop a simple and interpretable machine learning algorithm that first predicts the relevant lines in the document and then predicts the tumor attribute. Our results show across the small dataset sizes of 32, 64, 128, and 186 labeled documents per cancer, SLA only requires half the number of labeled documents as state-of-the-art methods to achieve similar or better micro-f1 and macro-f1 scores for the vast majority of comparisons that we made. Accounting for the increased annotation time, this leads to a 40\% reduction in total annotation time over the state of the art. 

\end{abstract}

\begin{keyword}
Pathology, Cancer, Natural Language Processing, Information Extraction
\end{keyword}
\end{frontmatter}

\section{Introduction}
\label{sec:intro}

 Precision medicine holds immense potential to improve the lives of patients and to fundamentally reimagine healthcare \citep{national2011toward} by enabling patients to receive tailored risk assessment and treatment decisions. Effective delivery of precision medicine depends on accurate and detailed patient data. Much of the relevant clinical data, such as tumor grade and histology for cancer patients, is stored as free text in lengthy unstructured or semi-structured reports \citep{burger_natural_2016}. Leveraging the data contained in these reports for precision medicine applications relies on manual efforts by annotators with domain expertise for many downstream automated methods. Due to the time-consuming and expensive nature of manual information extraction, researchers and clinicians have worked to develop algorithms to automatically extract pertinent data from pathology reports with mixed success, with machine learning-based methods underlying some of the more effective solutions \citep{burger_natural_2016}. However, generating a large training dataset for effective clinical natural language processing (NLP) is not always feasible due to the specific expertise required and strong privacy restrictions. This is especially true for oncology because there are many different cancer types that each has many relevant tumor attributes to extract. Furthermore, while there has been immense success applying deep learning to NLP in the last several years, these methods may not be suited for small clinical training dataset sizes, and in some cases, can substantially trail behind classical NLP methods \citep{odisho_pd58-09_2019}. Thus, it is critical to develop methods that can provide high accuracy using small training sets. 

Within clinical NLP in oncology, pathology reports in particular contain critical information for cancer patients, with details of the diagnosis and cellular analysis of the tumor tissue. This information is highly relevant for personalized treatment, risk assessment, clinical trial enrollment, and much more. Due to the clinical importance of these reports, previous work has attempted to classify tumor attributes using a variety of different machine learning methods, both classical and deep-learning. However, previous work has only given labels at the document level \citep{yala_using_2017,odisho_pd58-09_2019,gao_hierarchical_2018}. Due to the often long length of pathology reports, this leads to a ``needle in a haystack" problem for machine learning algorithms to determine which parts of the reports are relevant for the classification. The main insight of our work is that providing \textit{rationales} for the document level labels by highlighting relevant line or lines at annotation time can lead to large improvements in classifying tumor attributes. We describe a simple and interpretable machine learning method to leverage these enriched annotations for information extraction called Supervised Line Attention (SLA). We apply this method to classifying tumor attributes from pathology reports. We annotate 250 colon cancer pathology reports and 250 kidney cancer reports at the University of California, San Francisco and evaluate performance on training datasets with size 32, 64, 128, and 186 labeled documents. While we find that enriched annotations take around 20\% longer to generate than non-enriched annotations , we find that in a vast majority of cases SLA requires half the labeled data to achieve the same level of performance in terms of micro-f1 and macro-f1 as state-of-the-art methods, leading to a 40\% reduction in the annotation time required to achieve comparable performance. We therefore find that enriching annotations to encourage classifiers to be ``right for the right reasons" can greatly reduce the amount of labeled data needed and therefore how datasets are annotated can be as important as the machine learning algorithms themselves. 

The system that we have developed is in the process of being deployed within UCSF to be used by researchers and clinicians.

\section{Related Work}
The abundance of textual data in the clinical domain has led to increased interest in developing biomedical information extraction systems. These systems aim to automatically extract pre-specified data elements from medical documents, such as physician notes, radiology reports, and pathology reports, and store them in databases. Converting the originally free-text data into a structured form makes them easily available to clinical practitioners or researchers. 

For categorical attributes, the information extraction task can be viewed as an instance of document classification that classifies the tumor attribute based on document contents. For a given attribute, the value is one of a fixed set of options selected based on information in the document. As an illustration, the set of values for the attribute $``$presence of lymphovascular invasion$"$  could consist of the values $``$present$"$, $``$absent$"$,  and  $``$not reported$"$. Both classical and deep learning classification methods have been applied to this task in the prior work discussed below.

There has been success in applying classical machine learning techniques to classifying attributes of tumors from pathology reports. \citet{yala_using_2017} classified over 20 binary attributes from breast cancer pathology reports by training boosting classifiers over n-gram features using two corpuses consisting of 6,295 and 10,841 manually annotated reports. \citet{jouhet_automated_2012} investigated applications of Support Vector Machines (SVMs) and Naive Bayes classifiers to the task of predicting International Classification of Diseases for Oncology  (ICD-O-3) using 5125 cancer pathology reports total. More recently, there has been success in applying deep learning techniques to pathology report classification.  \citet{qiu_deep_2018} applied convolutional neural networks (CNNs) to predicting ICD-O-3 with 942 manually annotated breast and lung cancer pathology reports using 10-fold cross validation. \citet{gao_hierarchical_2018} applied hierarchical attention networks to predict tumor site using 645 labeled reports and grade using 942 labeled reports within the NCI-SEER dataset and noted improvement in micro-f1 averaged across 10 folds of up to 0.2 compared to baselines across primary site and histologic grade for lung and breast cancer.

There has also been work addressing pathology report classification in the absence of a large amount of labeled data. \citet{odisho_pd58-09_2019} analyzed performance of machine learning methods for extracting clinical information from prostate pathology reports across various data regimes and found that, while deep learning performed best when trained on the full dataset of 2,066 labeled documents and achieved a mean weighted-F1 score of 0.97 across classification attributes, simpler methods such as logistic regression and adaBoost performed best in smaller data regimes ($<$ 256 reports). Additionally, \citet{zhang-etal-2017-aspect} investigated the problem of unsupervised adaptation across attributes in breast cancer pathology reports. Given a set of attributes with labels and a new attribute without labels but with relevant keywords, they used adversarial adaptation with semi-supervised attention to extract data. We use all of  the above methods as baselines for our system to compare against, with the exception of  \citet{zhang-etal-2017-aspect} due to the difference in tasks.  

\section{Materials and Methods}
\subsection{Data}
\subsubsection*{Data Sources}
Our\ data consists of 250 colon cancer pathology reports and 250 kidney cancer reports from 2002-2019 at the University of California, San Francisco. The data was split into two sets, a set of 186 which we used for both training and validation and a test set of size 64. Institutional Review Board approval was obtained for this study.

\subsubsection*{Data Annotation Methods}
Pathology reports consist of free text describing a patient's clinical history and attributes describing the excised specimen, such as surgical procedure, cancer stage, tumor histology, grade, cell differentiation, and presence of invasion to surrounding tissues. More recent pathology reports also contain a synoptic comment section, which is a condensed semi-structured summary of relevant cancer attributes. While many of the most clinically important attributes are reported in this synoptic comment, this is not always the case. The presence and completeness of the synoptic comment also varies significantly over time. All attributes in the College of American Pathology reporting guidelines are annotated for each cancer, but for this paper we restrict our investigation to the most frequently occurring attributes in our pathology reports. These include tumor site, histologic type, procedure, laterality, tumor grade, and lymphovascular invasion for both cancers. Additionally, we have the cancer specific attributes of laterality and perineural invasion for kidney and colon cancer specifically.We list all attributes and their set of possible values in the appendix in Table~\ref{table:attributes}.  {The Multi-document Annotation Environment (MAE) \citet{stubbs_mae_2011} was used to annotate the documents. We do not consider the task of classifying cancer stage in this paper and instead treat it as an extraction task which we detail in the appendix.}

\subsubsection*{Enriched Annotations}
In previous work, annotations consisted of only the label for each attribute in a document \citep{yala_using_2017,odisho_pd58-09_2019,gao_hierarchical_2018}. However, in this work, for each attribute of interest the annotator highlighted all occurrences relevant to the label throughout the document, in addition to the label itself. This provides us with the specific location within the text that directly indicates the attribute's label. Each highlight is classified into the corresponding College of American Pathologists (CAP)-derived category. We investigate two types of annotation: the first we refer to as the ``minimal enriched annotation set", a minimal set of annotations containing the line of a given attribute value's first occurrence in the synoptic comment, or, if not in the synoptic comment, the line of where that information is referenced elsewhere in the document. The incremental time required to annotate this location is marginal because the annotator does not need to read any more of the document than that required to annotate the first occurrence of the attribute value. In fact, we investigated the amount of additional time required to create these enriched annotations and found that it took 20 percent longer on average, primarily due to the time it took the annotator to navigate the attribute drop-down menu. This could perhaps be improved through user interface considerations. In addition to a minimal enriched annotation set, we also investigate performance with all the occurrences relevant to the final classification highlighted, a more laborious annotation scheme. For our results, unless stated otherwise, we are using the minimal enriched annotation set due to its comparable annotation time to labeling the attribute values alone. 

If multiple labels for an attribute occurred within a report, we concatenate them to form a single composed label. For example, if the report contains both grade 1 and grade 2 as labels for histologic grade, we label the histologic grade of the report as $``$grade 1 and grade 2$"$. We provide additional data preprocessing in the appendix.

\subsection*{Baselines}
For all classical baselines, we represent each document as a union of a set of n-grams where n varies from 1 to N, where N is a hyperparameter. We consider values 1 to 4 as possible values for N. For all methods we use random search \citet{bergstra_random_2012} with 40 trials to tune our hyperparameters according to the 4-fold cross validation error which we found in preliminary experiments to be a good compromise between performance and computational efficiency. 
\subsubsection*{Logistic regression}

We use sklearn's \citet{pedregosa_scikit-learn:_2011} logistic regression model with L1 regularization and the liblinear solver. We use balanced class weights to up-weight the penalty on rare classes. We generate 500 points from -6 to 6 logspace for the regularization penalty, and sample 40 points at random.

\subsubsection*{Support Vector Classifier}

We use sklearn's SVC model with balanced class weights. We define our parameter space as 500 points evenly generated from -6 to 6 in log space for the error penalty \textit{C} of the model; the kernel as linear or rbf; and the parameter of the kernel as either 0.001, 0.01, 0.1, or 1. We then sample 40 points at random from this space.
\subsubsection*{Random Forest}
We use sklearn’s random forest classification model with balanced class weights. The parameter space consists of the number of estimators from 25, 50, 100, 200, 400, 600, 800, and 1000; the minimum number of samples for a leaf from 1 to 128 in powers of 2; max depth of a tree from 5, 10, 20, 30, 40, 50, 60, 70, 80, 90, and 100; and whether to bootstrap samples to build trees or not. We sample 40 points randomly from this parameter set.
    
\subsubsection*{Boosting}

We use sklearn's adaboost classifier with decision trees of depth 3 and with the SAMME.R boosting algorithm. Our parameter space is 500 points generated evenly from -4 to 1 in logspace for the learning rate and either 25, 50, 100, 25, or 500 for the number of estimators. We then sample 40 points at random from this space.

\subsubsection*{Hierarchical Attention Network}

We implement the hierarchical attention method from 
\citet{gao_hierarchical_2018} which was shown to be state-of-the-art for classifying tumor site and grade from lung cancer and breast cancer pathology reports within the NCI-SEER dataset. This model represents the document as a series of word-vectors. For each sentence in the document it runs a gated recurrent unit (GRU) \citet{cho_learning_2014} over the word vectors. It then uses an attention module to create a sentence representation as a sum of the attention-weighted outputs of the GRU. To generate the document representation, a GRU is run over the sentence representations, followed by another attention module applied to the GRU outputs. The document representation is the attention-weighted sum of the GRU outputs. Since it was detrimental to performance as often as it was beneficial, we did not experiment with the unsupervised pre-training method in \citep{gao_hierarchical_2018}.
    
For our hyperparameters we use random search across the learning rate, which is either 1e-2, 1e-3, or 1e-4; the width of the hidden layer of the attention module, which is either 50, 100, 150, 200, 250, or 500; the hidden size of the GRU, which is either 50, 100, 150, 200, 250, or 500; and the dropout rate applied to the document representation, which is either 0, 0.2, 0.4, 0.6, or 0.8. We use a batch size of 64 and ADAM \citet{kingma_adam_2014-1} as our optimizer.

\subsection{Our method: Supervised Line Attention}

In order to take advantage of annotations enriched with location data, we propose a two-stage prediction procedure in which we first predict which lines in the document contain relevant information. We then concatenate the predicted relevant lines and use this string to make the final class prediction using logistic regression.

\subsubsection*{Finding Relevant Lines}

The first stage predicts which lines are relevant to the attribute. We do this by training an xgboost \citet{chen2016xgboost} binary classification model that takes a line represented as a bag of n-grams as its input and outputs whether or not the line is relevant to the attribute. The relevance of each line is predicted independently by this initial classifier.

We then take the top-k lines with the highest scores under the model (where k is a hyperparameter). Groups of adjacent lines are conjoined into one line so that sentences which span multiple lines are presented to the model as a single line.

Finally, we represent each line as a set of n-grams vectors and compose a document representation as the weighted sum of each vector representation, which is weighted by the score of that line under the xgboost model. If a line is conjoined, its weight is the maximum of all the xgboost scores for each line in the conjoined line. Mathematically, this is represented as 

\begin{center}
 \[ d_{r} \left( l_{1},...,l_{n} \right)  =  \sum _{l_{i}  \epsilon  S_{k}}^{}v \left( l_{i} \right) m \left( l_{i} \right)  \] 
\end{center}\par

where  \( d_{r} \) represents the vector representation of a document  \( d \) , \( S_{k} \)  are the top-k lines with the highest scores under the xgboost model, \( v \) is the mapping from a line  \( l_{i} \) to its set of n-grams representation, and  \( m \left( l_{i} \right)  \)  is the xgboost score for line  \( l_{i} \) .

With this final weighted representation, we train an L1 regularized logistic regression model with balanced class weights to predict the final class.

We refer to this method as $``$supervised line attention$"$  due to its relationship to supervised attention in the deep learning literature which predicts relevant locations and creates a weighted representation of the relevant regions' features. Supervised attention in the deep learning literature has been used to match a neural machine translations attention distribution an unsupervised aligner \citet{liu_neural_2016} and to match a sequence-to-sequence neural constituency parser's attention mechanism with traditional parsing features \citet{kamigaito_supervised_2017}, for example. Our approach can be viewed as a form of supervised attention for document classification. The principle difference from existing work is that in supervised attention in the deep learning literature, the method is trained in an end-to-end fashion with neural networks, whereas we train each module independently with classical methods and our feature representation for sentences are sets of n-grams instead of dense real-valued vectors.

\subsubsection*{Rule-based line classifier}

As a baseline, we also include a line classifier that selects relevant lines by searching for expert-generated keywords and phrases. After the lines are selected, the final representation is generated the same way, with the exception that all selected lines are given a weight of 1; thus, for all  \( l_{i~} \epsilon~  S_{k} \) ,  \( m \left( l_{i} \right)  = 1. \)

\subsubsection*{Oracle Model}

In addition to the line attention model, we also evaluate a model\textit{ }that uses the correct relevant lines from the annotator directly as input to the final classifier, which we refer to as the $``$oracle model$"$. Using the oracle lines, the final representation is generated the same way as the rule-based line classifier, where all lines are given a weight of 1.

\subsubsection*{Hyperparameter tuning}

Similar to our baselines, we perform random search for 40 iterations and choose the hyperparameters that minimize 4-fold cross-validation error. The hyperparameters for our shallow attention method are an n-gram size for finding relevant lines between 1 and 4; an n-gram size for the second stage of making the final classification between 1 to 4.

For \textit{xgboost}, the hyperparameters were 500 points from -2 to -0.5 in logspace for the learning rate; a max depth between 3 to 7; a minimum split loss reduction to split a node that is 0, 0.01, 0.05, 0.1, 0.5, or 1; a subsample ratio that is 0.5, 0.75, or 1; and an L2 regularization on the weights that is 0.1, 0.5, 1, 1.5, or 2.

For the final classifier, the L1 penalty is chosen from 500 evenly spaced points from -6 to 6 in logspace. Additionally, since the final representation is a weighted representation of the features of the top-k lines under the line classifier model, we have a hyperparameter \textit{k} which determines how many lines to use, where \textit{k} is between 1 and 5.

\begin{figure}
	
	\centering
		\includegraphics[width=\textwidth]{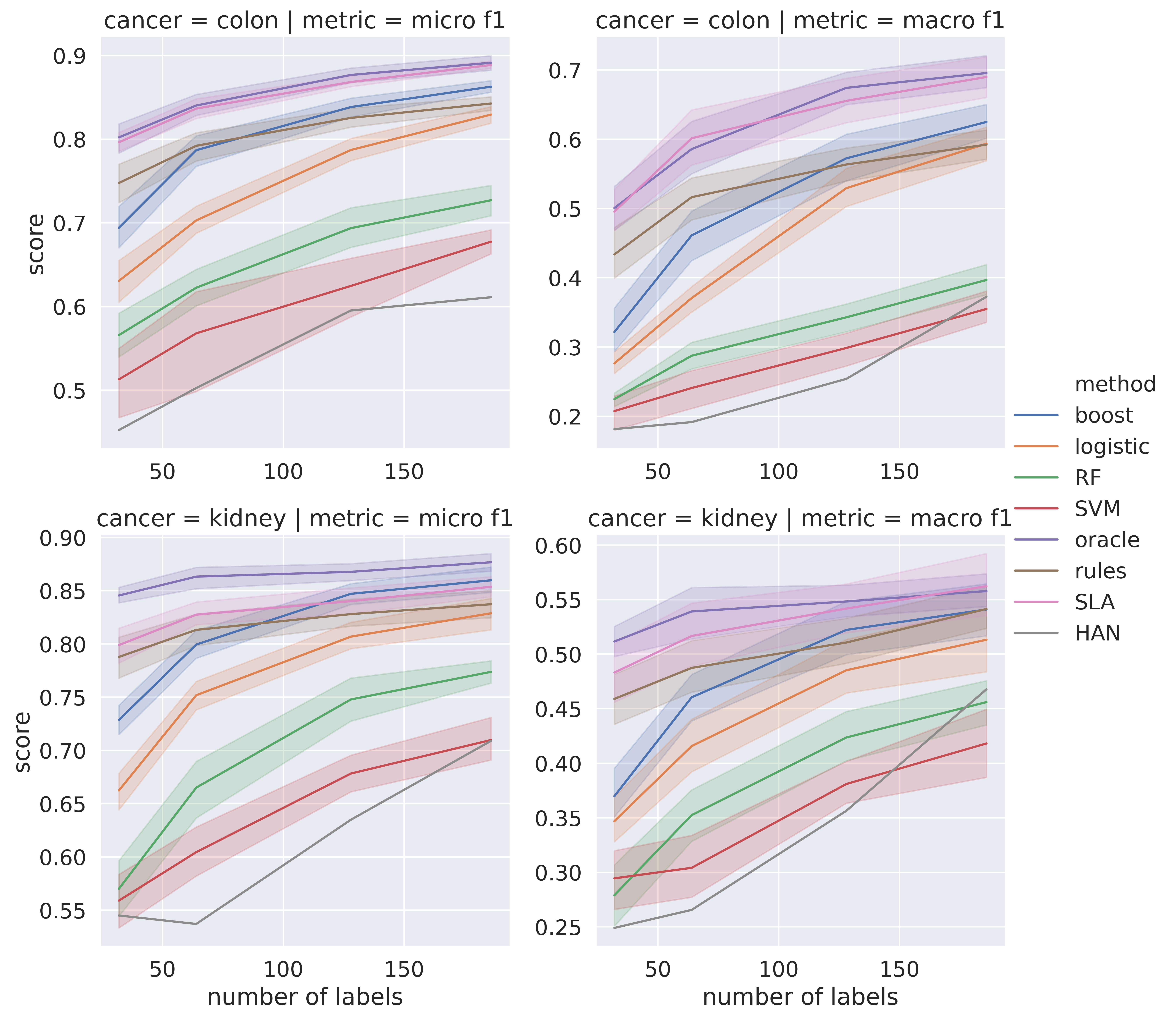}
	\caption[width=.5\textwidth]{Average micro-f1 and macro-f1 performance across attributes of methods as a function of the number labeled colon and kidney cancer pathology reports. SLA: supervised line attention; oracle: oracle model that gets access to the true lines as input; rules: line prediction done with a rule-based method and logistic regression to predict the final class; boost: XGBoost; SVM: Support Vector Machine; logistic; logistic regression; RF: Random forest; HAN: Hierarchical attention network. We see that SLA outperforms existing methods in almost all cases.}
	
	\label{fig:main_results}

\end{figure}

\section{Results}

We trained our methods using various training set sizes of 32, 64, 128, and 186 with 4-fold cross validation. We take the average of 10 runs where we reshuffle the data and generate new splits each time and compute 95$\%$  confidence intervals for all methods using bootstrap resampling, with the exception of the HAN method due to computational limitations. For our results, unless stated otherwise, we are using the minimal enriched annotation set due to its comparable annotation time to labeling\ the\ attribute values alone. As shown in Figure~\ref{fig:main_results},  our supervised line attention model frequently improves substantially over existing methods in terms of micro and macro-f1, particularly in the lowest data regimes. For example, for colon cancer we see an absolute improvement of 0.10 micro-f1 and 0.17 macro-f1 over previously existing methods with 32 labeled data points. Furthermore, SLA frequently tends to perform as well or better than state-of-the-art methods with only half the labeled documents. Two exceptions are in kidney cancer micro-f1 scores, where boosting performs .01 better in micro-f1 for 128 and 186 labeled documents. We see that the rule-based line classifier method tends to do better than existing methods with 64 labeled data points or fewer, but its performance plateaus and XGBoost outperforms it with 128 and 186 labeled data points. Furthermore, we see that  the rule-based line classifier consistently performs worse than supervised line attention. We provide a detailed table of the performance of different methods across data regimes and cancers in Table~\ref{table:results_table} and ablations of our method in the appendix.

\subsection{Minimal Enriched Annotations vs Full Enriched Annotations}

\begin{figure}
	
	\centering
		\includegraphics[width=\textwidth]{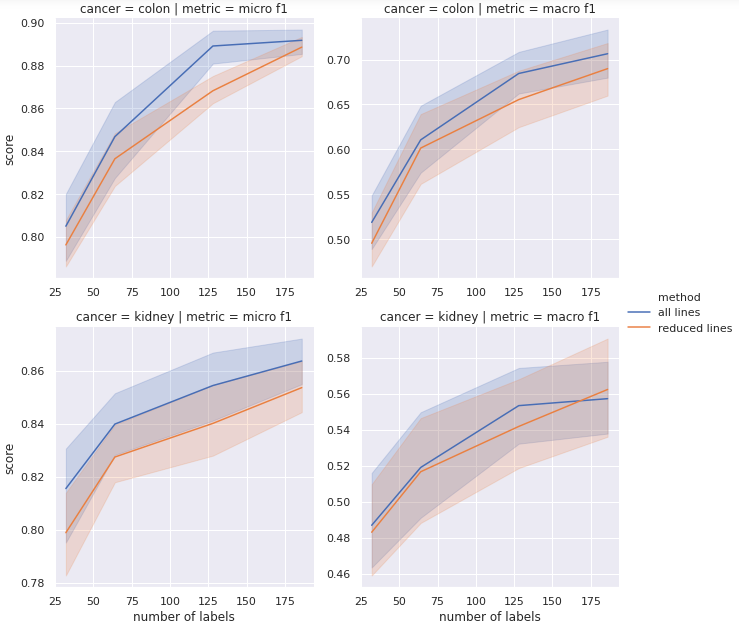}
	\caption[width=.5\textwidth]{Comparing the more laborious scheme of annotating the location information for all relevant lines for a given attribute as compared to the more lightweight annotation method of only annotating the first line in the synoptic comment, if the synoptic comment contains the information, or the first relevant line in the document otherwise. We see that having the additional information yields a consistent, though sometimes small, benefit.}
	
	\label{fig:reduced}

\end{figure}

Here we compare how well reduced annotation compares to the more laborious full enriched annotation setting where an annotator highlights all lines in the document relevant to final classification. We use the same setup as our main results and present our results in Figure~\ref{fig:reduced}. We can see that the full enriched annotation set leads to a consistent increase in performance. However, it is unclear whether the extra time required to create this full enriched annotation scheme is beneficial overall as it would lead to fewer documents annotated in the same amount of time.




\subsection{Error Analysis}

To better understand model performance, we inspect all errors that the supervised line attention model makes for each attribute and cancer domain. In our investigation we find 6 primary types of errors, which we define below:

An \textit{attribute qualification error} occurs when the model correctly extracts the relevant lines, but fails to classify the final label correctly because the label text is negated or qualified by an additional phrase indicating information is not available, such as in the following example: ``If we were to classify the tumor, it would be grade 2 but due to the treatment effect it is unclassified."
A \textit{rare phrasing error} occurs when the model correctly predicts the relevant lines, but the relevant lines contain rare or unusual phrasing and the model assigns an incorrect final classification.
An \textit{irrelevant lines error} occurs when the model includes irrelevant lines in its final predictions, which can influence the final classification.
A \textit{multi-label error} occurs when a report contains a conjoined label (such as $``$grade 1 and grade 2$"$ ), but the model only correctly predicts one of the labels.
An \textit{annotator error} occurs when the model's prediction is correct, but on re-review we noted that the annotator's label was incorrect.
An \textit{unknown error} occurs when the underlying cause of the error is not known. This often occurs when the model correctly extracts out the relevant line but assigns an incorrect final label. 

We found that the most common error is the multi-label error accounting for 24/79 of the errors our model made. This is primarily problematic for colon cancer histologic grade, where pathologists will describe a range of grades such as $``$grade 1-2$"$  and tumor site for colon and kidney cancer as tumors can inhabit multiple sites. This suggests that treating this as a multi-label classification problem instead of naively conjoining multiple labels may reduce many of the errors. We note that annotator errors account for 15/79 of our errors, suggesting that higher quality annotations may also lead to an improvement in performance. We provide a detailed breakdown of the different errors across cancers and attributes in Table~\ref{table:err_colon} and Table~\ref{table:err_kidney} in the appendix.
\section{Discussion}

We have investigated the efficacy of location-enriched annotations and a corresponding simple and interpretable method, which we call Supervised Line Attention (SLA), for extracting data elements from pathology reports across colon and kidney cancers at UCSF. By leveraging location annotations, our two-stage modeling approach can lead to increases of micro-f1 scores up to 0.1 and macro-f1 scores up to 0.17 over state-of-the-art methods and typically reduces the required annotation by 40\% to achieve the performance of existing methods.

Our SLA approach with enriched annotations was primarily developed to tackle the problem of achieving accurate performance with minimal labeled data. Previous approaches that attempt to leverage additional data use multi-task learning and transfer learning using information from other cancer domains with complex modeling architectures. For example, \citet{qiu_deep_2018} investigated using transfer learning with convolutional neural networks to extract data from 942 breast and lung cancer reports, achieving 0.685 and 0.782 micro-f1 scores, respectively. \citet{10.1093/jamia/ocz153} implemented multitask learning with convolutional neural networks to classify tumor attributes in 942 pathology reports for breast and lung cancers, and achieved 0.77, 0.79, and 0.96 micro-f1 scores for tumor site, histologic grade, and laterality, respectively. While, we use different datasets, we do note here that with a much smaller dataset we see similar micro-f1 scores of .89 for lung and .85 for kidney cancer.

An important observation is that our approach is more interpretable than previous machine learning methods, since addition to outputting the probability and predicted value for a certain report, our system outputs the exact lines of the text used to make the classification as well. This enables practitioners to easily check predictions by examining the lines output by the extraction system, and verify the system is working as expected before making clinical decisions. The hierarchical attention approach used by \citet{gao_hierarchical_2018} also can output the most pertinent sentences for a classification by using the attention mechanism to hierarchically filter out pieces of text. However, our experiments show that HAN requires a large training size to achieve adequate performance due to the more complex architecture used, and requires significantly more development and computational time to search the hyperparameter space. Additionally, there have been recent concerns regarding the  interpretability of attention distributions from neural networks \citep{jain_attention_2019}.

    Our study has a few limitations. Although we observe high performance of our methodology in both colon and kidney cancer reports at UCSF, our investigation was done at a single institution; this may limit the generalizability of our findings to other institutions that use different pathology reporting or data collection systems. Second, within the field of natural language processing, there has been strong empirical evidence showing the benefit of pre-trained contextualized representations for a variety of tasks, both in and out of clinical applications \citep{peters_deep_2018,devlin_bert_2018,lee_biobert_2019-1,huang_clinicalbert_2019-1}. In preliminary experiments, we investigated the efficacy of using biomedical word vectors \citet{pyysalo_distributional_2013} as feature representation input to our SLA model, but did not see an improvement in results. However, it would be interesting to investigate the effect that more sophisticated contextualized representations may have on downstream performance, and this may increase performance of SLA even further.

\section{Conclusion}
To advance the application of information extraction in medicine, where labeled data is scarce and expensive to acquire, we have developed a novel annotation scheme and simple and interpretable classification method, which can reduce the annotation time by 40\% for accurate tumor attribute classification compared to state-of-the-art approaches.  By outputting the exact lines in the text used to determine the final classification, our supervised line attention method allows for greater interpretability, which can allow for easy verification of its outputs for clinicians.

\acks
Partial support is gratefully acknowledged from ARO grant W911NF1710005, NSF grants DMS-1613002 and IIS 1741340, the Center for Science of Information (CSoI), a US NSF Science and Technology Center, under grant agreement CCF-0939370, Adobe Research Grant, and the Bakar Computational Health Sciences Institute, University of California, San Francisco.


\bibliographystyle{elsarticle-num-names}
\bibliography{sample}

\begin{thebibliography}{23}
\expandafter\ifx\csname natexlab\endcsname\relax\def\natexlab#1{#1}\fi
\providecommand{\url}[1]{\texttt{#1}}
\providecommand{\href}[2]{#2}
\providecommand{\path}[1]{#1}
\providecommand{\DOIprefix}{doi:}
\providecommand{\ArXivprefix}{arXiv:}
\providecommand{\URLprefix}{URL: }
\providecommand{\Pubmedprefix}{pmid:}
\providecommand{\doi}[1]{\href{http://dx.doi.org/#1}{\path{#1}}}
\providecommand{\Pubmed}[1]{\href{pmid:#1}{\path{#1}}}
\providecommand{\bibinfo}[2]{#2}
\ifx\xfnm\relax \def\xfnm[#1]{\unskip,\space#1}\fi
\bibitem[{Council et~al.(2011)}]{national2011toward}
\bibinfo{author}{N.~R. Council}, et~al., \bibinfo{title}{Toward precision
  medicine: building a knowledge network for biomedical research and a new
  taxonomy of disease}, \bibinfo{publisher}{National Academies Press},
  \bibinfo{year}{2011}.
\bibitem[{Burger et~al.(2016)Burger, Abu-Hanna, de~Keizer, and
  Cornet}]{burger_natural_2016}
\bibinfo{author}{G.~Burger}, \bibinfo{author}{A.~Abu-Hanna},
  \bibinfo{author}{N.~de~Keizer}, \bibinfo{author}{R.~Cornet},
\newblock \bibinfo{title}{Natural language processing in pathology: a scoping
  review},
\newblock \bibinfo{journal}{Journal of Clinical Pathology}
  (\bibinfo{year}{2016}). \DOIprefix\doi{10.1136/jclinpath-2016-203872}.
\bibitem[{Odisho* et~al.(2019)Odisho*, Park, Altieri, Murdoch, Carroll,
  Coopberberg, and Yu}]{odisho_pd58-09_2019}
\bibinfo{author}{A.~Odisho*}, \bibinfo{author}{B.~Park},
  \bibinfo{author}{N.~Altieri}, \bibinfo{author}{W.~Murdoch},
  \bibinfo{author}{P.~Carroll}, \bibinfo{author}{M.~Coopberberg},
  \bibinfo{author}{B.~Yu},
\newblock \bibinfo{title}{{PD58}-09 {Extracting} {Structured} {Information}
  {From} {Pathology} {Reports} {Using} {Natural} {Language} {Processing} {And}
  {Machine} {Learning}},
\newblock \bibinfo{journal}{The Journal of Urology}  (\bibinfo{year}{2019}).
  \URLprefix
  \url{https://www.auajournals.org/doi/10.1097/01.JU.0000557177.97226.63}.
\bibitem[{Yala et~al.(2017)Yala, Barzilay, Salama, Griffin, Sollender, Bardia,
  Lehman, Buckley, Coopey, Polubriaginof, Garber, Smith, Gadd, Specht,
  Gudewicz, Guidi, Taghian, and Hughes}]{yala_using_2017}
\bibinfo{author}{A.~Yala}, \bibinfo{author}{R.~Barzilay},
  \bibinfo{author}{L.~Salama}, \bibinfo{author}{M.~Griffin},
  \bibinfo{author}{G.~Sollender}, \bibinfo{author}{A.~Bardia},
  \bibinfo{author}{C.~Lehman}, \bibinfo{author}{J.~M. Buckley},
  \bibinfo{author}{S.~B. Coopey}, \bibinfo{author}{F.~Polubriaginof},
  \bibinfo{author}{J.~E. Garber}, \bibinfo{author}{B.~L. Smith},
  \bibinfo{author}{M.~A. Gadd}, \bibinfo{author}{M.~C. Specht},
  \bibinfo{author}{T.~M. Gudewicz}, \bibinfo{author}{A.~J. Guidi},
  \bibinfo{author}{A.~Taghian}, \bibinfo{author}{K.~S. Hughes},
\newblock \bibinfo{title}{Using machine learning to parse breast pathology
  reports},
\newblock \bibinfo{journal}{Breast Cancer Research and Treatment}
  \bibinfo{volume}{161} (\bibinfo{year}{2017}) \bibinfo{pages}{203--211}.
  \DOIprefix\doi{10.1007/s10549-016-4035-1}.
\bibitem[{Gao et~al.(2018)Gao, Young, Qiu, Yoon, Christian, Fearn, Tourassi,
  and Ramanthan}]{gao_hierarchical_2018}
\bibinfo{author}{S.~Gao}, \bibinfo{author}{M.~T. Young}, \bibinfo{author}{J.~X.
  Qiu}, \bibinfo{author}{H.-J. Yoon}, \bibinfo{author}{J.~B. Christian},
  \bibinfo{author}{P.~A. Fearn}, \bibinfo{author}{G.~D. Tourassi},
  \bibinfo{author}{A.~Ramanthan},
\newblock \bibinfo{title}{Hierarchical attention networks for information
  extraction from cancer pathology reports},
\newblock \bibinfo{journal}{Journal of the American Medical Informatics
  Association} \bibinfo{volume}{25} (\bibinfo{year}{2018})
  \bibinfo{pages}{321--330}. \URLprefix
  \url{https://academic.oup.com/jamia/article/25/3/321/4636780}.
  \DOIprefix\doi{10.1093/jamia/ocx131}.
\bibitem[{Jouhet et~al.(2012)Jouhet, Defossez, Burgun, le~Beux, Levillain,
  Ingrand, and Claveau}]{jouhet_automated_2012}
\bibinfo{author}{V.~Jouhet}, \bibinfo{author}{G.~Defossez},
  \bibinfo{author}{A.~Burgun}, \bibinfo{author}{P.~le~Beux},
  \bibinfo{author}{P.~Levillain}, \bibinfo{author}{P.~Ingrand},
  \bibinfo{author}{V.~Claveau},
\newblock \bibinfo{title}{Automated classification of free-text pathology
  reports for registration of incident cases of cancer},
\newblock \bibinfo{journal}{Methods of Information in Medicine}
  \bibinfo{volume}{51} (\bibinfo{year}{2012}) \bibinfo{pages}{242--251}.
  \DOIprefix\doi{10.3414/ME11-01-0005}.
\bibitem[{Qiu et~al.(2018)Qiu, Yoon, Fearn, and Tourassi}]{qiu_deep_2018}
\bibinfo{author}{J.~X. Qiu}, \bibinfo{author}{H.-J. Yoon},
  \bibinfo{author}{P.~A. Fearn}, \bibinfo{author}{G.~D. Tourassi},
\newblock \bibinfo{title}{Deep {Learning} for {Automated} {Extraction} of
  {Primary} {Sites} {From} {Cancer} {Pathology} {Reports}},
\newblock \bibinfo{journal}{IEEE journal of biomedical and health informatics}
  \bibinfo{volume}{22} (\bibinfo{year}{2018}) \bibinfo{pages}{244--251}.
  \DOIprefix\doi{10.1109/JBHI.2017.2700722}.
\bibitem[{Zhang et~al.(2017)Zhang, Barzilay, and
  Jaakkola}]{zhang-etal-2017-aspect}
\bibinfo{author}{Y.~Zhang}, \bibinfo{author}{R.~Barzilay},
  \bibinfo{author}{T.~Jaakkola},
\newblock \bibinfo{title}{Aspect-augmented adversarial networks for domain
  adaptation},
\newblock \bibinfo{journal}{Transactions of the Association for Computational
  Linguistics}  (\bibinfo{year}{2017}). \URLprefix
  \url{https://www.aclweb.org/anthology/Q17-1036}.
\bibitem[{Stubbs(2011)}]{stubbs_mae_2011}
\bibinfo{author}{A.~Stubbs},
\newblock \bibinfo{title}{{MAE} and {MAI}: {Lightweight} {Annotation} and
  {Adjudication} {Tools}},
\newblock \bibinfo{year}{2011}, pp. \bibinfo{pages}{129--133}. \URLprefix
  \url{https://www.aclweb.org/anthology/W11-0416}.
\bibitem[{Bergstra and Bengio(2012)}]{bergstra_random_2012}
\bibinfo{author}{J.~Bergstra}, \bibinfo{author}{Y.~Bengio},
\newblock \bibinfo{title}{Random search for hyper-parameter optimization},
\newblock \bibinfo{journal}{Journal of Machine Learning Research}
  (\bibinfo{year}{2012}) \bibinfo{pages}{281--305}.
\bibitem[{Pedregosa et~al.(2011)Pedregosa, Varoquaux, Gramfort, Michel,
  Thirion, Grisel, Blondel, Prettenhofer, Weiss, Dubourg, and
  {others}}]{pedregosa_scikit-learn:_2011}
\bibinfo{author}{F.~Pedregosa}, \bibinfo{author}{G.~Varoquaux},
  \bibinfo{author}{A.~Gramfort}, \bibinfo{author}{V.~Michel},
  \bibinfo{author}{B.~Thirion}, \bibinfo{author}{O.~Grisel},
  \bibinfo{author}{M.~Blondel}, \bibinfo{author}{P.~Prettenhofer},
  \bibinfo{author}{R.~Weiss}, \bibinfo{author}{V.~Dubourg},
  \bibinfo{author}{{others}},
\newblock \bibinfo{title}{Scikit-learn: {Machine} learning in {Python}},
\newblock \bibinfo{journal}{Journal of machine learning research}
  \bibinfo{volume}{12} (\bibinfo{year}{2011}) \bibinfo{pages}{2825--2830}.
\bibitem[{Cho et~al.(2014)Cho, Merriënboer, Gulcehre, Bahdanau, Bougares,
  Schwenk, and Bengio}]{cho_learning_2014}
\bibinfo{author}{K.~Cho}, \bibinfo{author}{B.~v. Merriënboer},
  \bibinfo{author}{C.~Gulcehre}, \bibinfo{author}{D.~Bahdanau},
  \bibinfo{author}{F.~Bougares}, \bibinfo{author}{H.~Schwenk},
  \bibinfo{author}{Y.~Bengio},
\newblock \bibinfo{title}{Learning {Phrase} {Representations} using {RNN}
  {Encoder}–{Decoder} for {Statistical} {Machine} {Translation}},
\newblock \bibinfo{year}{2014}, pp. \bibinfo{pages}{1724--1734}. \URLprefix
  \url{https://www.aclweb.org/anthology/D14-1179}.
  \DOIprefix\doi{10.3115/v1/D14-1179}.
\bibitem[{Kingma and Ba(2014)}]{kingma_adam_2014-1}
\bibinfo{author}{D.~P. Kingma}, \bibinfo{author}{J.~Ba},
\newblock \bibinfo{title}{Adam: {A} {Method} for {Stochastic} {Optimization}}
  (\bibinfo{year}{2014}). \URLprefix \url{https://arxiv.org/abs/1412.6980v9}.
\bibitem[{Chen and Guestrin(2016)}]{chen2016xgboost}
\bibinfo{author}{T.~Chen}, \bibinfo{author}{C.~Guestrin},
\newblock \bibinfo{title}{Xgboost: A scalable tree boosting system},
\newblock in: \bibinfo{booktitle}{Proceedings of the 22nd acm sigkdd
  international conference on knowledge discovery and data mining},
  \bibinfo{year}{2016}, pp. \bibinfo{pages}{785--794}.
\bibitem[{Liu et~al.(2016)Liu, Utiyama, Finch, and Sumita}]{liu_neural_2016}
\bibinfo{author}{L.~Liu}, \bibinfo{author}{M.~Utiyama},
  \bibinfo{author}{A.~Finch}, \bibinfo{author}{E.~Sumita},
\newblock \bibinfo{title}{Neural {Machine} {Translation} with {Supervised}
  {Attention}},
\newblock \bibinfo{year}{2016}, pp. \bibinfo{pages}{3093--3102}. \URLprefix
  \url{https://www.aclweb.org/anthology/C16-1291}.
\bibitem[{Kamigaito et~al.(2017)Kamigaito, Hayashi, Hirao, Takamura, Okumura,
  and Nagata}]{kamigaito_supervised_2017}
\bibinfo{author}{H.~Kamigaito}, \bibinfo{author}{K.~Hayashi},
  \bibinfo{author}{T.~Hirao}, \bibinfo{author}{H.~Takamura},
  \bibinfo{author}{M.~Okumura}, \bibinfo{author}{M.~Nagata},
\newblock \bibinfo{title}{Supervised {Attention} for {Sequence}-to-{Sequence}
  {Constituency} {Parsing}},
\newblock \bibinfo{year}{2017}, pp. \bibinfo{pages}{7--12}. \URLprefix
  \url{https://www.aclweb.org/anthology/I17-2002}.
\bibitem[{Alawad et~al.(2019)Alawad, Gao, Qiu, Yoon, Blair~Christian,
  Penberthy, Mumphrey, Wu, Coyle, and Tourassi}]{10.1093/jamia/ocz153}
\bibinfo{author}{M.~Alawad}, \bibinfo{author}{S.~Gao}, \bibinfo{author}{J.~X.
  Qiu}, \bibinfo{author}{H.~J. Yoon}, \bibinfo{author}{J.~Blair~Christian},
  \bibinfo{author}{L.~Penberthy}, \bibinfo{author}{B.~Mumphrey},
  \bibinfo{author}{X.-C. Wu}, \bibinfo{author}{L.~Coyle},
  \bibinfo{author}{G.~Tourassi},
\newblock \bibinfo{title}{{Automatic extraction of cancer registry reportable
  information from free-text pathology reports using multitask convolutional
  neural networks}},
\newblock \bibinfo{journal}{Journal of the American Medical Informatics
  Association} \bibinfo{volume}{27} (\bibinfo{year}{2019})
  \bibinfo{pages}{89--98}. \URLprefix
  \url{https://doi.org/10.1093/jamia/ocz153}.
  \DOIprefix\doi{10.1093/jamia/ocz153}.
\bibitem[{Jain and Wallace(2019)}]{jain_attention_2019}
\bibinfo{author}{S.~Jain}, \bibinfo{author}{B.~C. Wallace},
\newblock \bibinfo{title}{Attention is not {Explanation}},
\newblock \bibinfo{journal}{arXiv:1902.10186 [cs]}  (\bibinfo{year}{2019}).
  \URLprefix \url{http://arxiv.org/abs/1902.10186}, \bibinfo{note}{arXiv:
  1902.10186}.
\bibitem[{Peters et~al.(2018)Peters, Neumann, Iyyer, Gardner, Clark, Lee, and
  Zettlemoyer}]{peters_deep_2018}
\bibinfo{author}{M.~E. Peters}, \bibinfo{author}{M.~Neumann},
  \bibinfo{author}{M.~Iyyer}, \bibinfo{author}{M.~Gardner},
  \bibinfo{author}{C.~Clark}, \bibinfo{author}{K.~Lee},
  \bibinfo{author}{L.~Zettlemoyer},
\newblock \bibinfo{title}{Deep contextualized word representations}
  (\bibinfo{year}{2018}). \URLprefix \url{https://arxiv.org/abs/1802.05365v2}.
\bibitem[{Devlin et~al.(2018)Devlin, Chang, Lee, and
  Toutanova}]{devlin_bert_2018}
\bibinfo{author}{J.~Devlin}, \bibinfo{author}{M.-W. Chang},
  \bibinfo{author}{K.~Lee}, \bibinfo{author}{K.~Toutanova},
\newblock \bibinfo{title}{{BERT}: {Pre}-training of {Deep} {Bidirectional}
  {Transformers} for {Language} {Understanding}}  (\bibinfo{year}{2018}).
  \URLprefix \url{https://arxiv.org/abs/1810.04805v2}.
\bibitem[{Lee et~al.(2019)Lee, Yoon, Kim, Kim, Kim, So, and
  Kang}]{lee_biobert_2019-1}
\bibinfo{author}{J.~Lee}, \bibinfo{author}{W.~Yoon}, \bibinfo{author}{S.~Kim},
  \bibinfo{author}{D.~Kim}, \bibinfo{author}{S.~Kim}, \bibinfo{author}{C.~H.
  So}, \bibinfo{author}{J.~Kang},
\newblock \bibinfo{title}{{BioBERT}: a pre-trained biomedical language
  representation model for biomedical text mining}  (\bibinfo{year}{2019}).
  \URLprefix \url{https://arxiv.org/abs/1901.08746v4}.
  \DOIprefix\doi{10.1093/bioinformatics/btz682}.
\bibitem[{Huang et~al.(2019)Huang, Altosaar, and
  Ranganath}]{huang_clinicalbert_2019-1}
\bibinfo{author}{K.~Huang}, \bibinfo{author}{J.~Altosaar},
  \bibinfo{author}{R.~Ranganath},
\newblock \bibinfo{title}{{ClinicalBERT}: {Modeling} {Clinical} {Notes} and
  {Predicting} {Hospital} {Readmission}}  (\bibinfo{year}{2019}). \URLprefix
  \url{https://arxiv.org/abs/1904.05342v2}.
\bibitem[{Pyysalo et~al.(2013)Pyysalo, Ginter, Moen, Salakoski, and
  Ananiadou}]{pyysalo_distributional_2013}
\bibinfo{author}{S.~Pyysalo}, \bibinfo{author}{F.~Ginter},
  \bibinfo{author}{H.~Moen}, \bibinfo{author}{T.~Salakoski},
  \bibinfo{author}{S.~Ananiadou}, \bibinfo{title}{Distributional {Semantics}
  {Resources} for {Biomedical} {Text} {Processing}}, \bibinfo{year}{2013}.
  \bibinfo{note}{Library Catalog: www.semanticscholar.org}.

\end{thebibliography}

\appendix

\section{First Appendix}\label{apd:first}











\begin{longtable}{p{1.0000in}p{5.3000in}}
\caption{Extracted attributes and their possible values}
\label{table:attributes}
\endfirsthead
\multicolumn{2}{c}{\textit{continued from previous page }} 
\endhead
\multicolumn{2}{r}{\textit{continued on next page}} \\
\endfoot
\endlastfoot\hline
\hline
\multicolumn{1}{|p{1.0000in}}{Tumor Site} & 
\multicolumn{1}{p{5.3000in}|}{} \\
\hhline{--}
\multicolumn{1}{|p{1.0000in}}{Colon} & 
\multicolumn{1}{|p{5.3000in}|}{Cannot be determined, cecum, colon not otherwise specified, hepatic flexure, ileocecal valve, left descending colon, other, rectosigmoid junction, rectum, right ascending colon, sigmoid colon, splenic flexure, transverse colon, or not reported} \\
\hhline{--}
\multicolumn{1}{|p{1.0000in}}{Kidney} & 
\multicolumn{1}{|p{5.3000in}|}{Upper pole, middle pole, lower pole, other, not specified, or not reported} \\
\hhline{--}
\multicolumn{1}{|p{1.0000in}}{Histologic Type} & 
\multicolumn{1}{p{5.3000in}|}{} \\
\hhline{--}
\multicolumn{1}{|p{1.0000in}}{Colon} & 
\multicolumn{1}{|p{5.3000in}|}{Adenocarcinoma, adenosquamous carcinoma, carcinoma, type cannot be determined, large cell neuroendocrine carcinoma, medullary carcinoma, micropapillary carcinoma, mucinous adenocarcinoma, neuroendocrine carcinoma poorly differentiated, other histologic type not listed, serrated adenocarcinoma, signet-ring cell carcinoma, small cell neuroendocrine carcinoma, squamous cell carcinoma, undifferentiated carcinoma, or not reported} \\
\hhline{--}
\multicolumn{1}{|p{1.0000in}}{Kidney} & 
\multicolumn{1}{|p{5.3000in}|}{Acquired cystic disease associated renal cell carcinoma, chromophobe renal cell carcinoma, clear cell papillary renal cell carcinoma, clear cell renal cell carcinoma, collecting duct carcinoma, hereditary leiomyomatosis and renal cell carcinoma-associated renal cell carcinoma, mit family translocation renal cell carcinoma, mucinous tubular and spindle renal cell carcinoma, multilocular cystic clear cell renal cell neoplasm of low malignant potential, oncocytoma, other histologic type, papillary renal cell carcinoma, papillary renal cell carcinoma type 1, papillary renal cell carcinoma type 2, renal cell carcinoma unclassified, renal medullary carcinoma, succinate dehydrogenase sdh deficient renal cell carcinoma, t611 renal cell carcinoma, tubulocystic renal cell carcinoma, xp11 translocation renal cell carcinoma, or not reported} \\
\hhline{--}
\multicolumn{1}{|p{1.0000in}}{Procedure} & 
\multicolumn{1}{p{5.3000in}|}{} \\
\hhline{--}
\multicolumn{1}{|p{1.0000in}}{Colon} & 
\multicolumn{1}{|p{5.3000in}|}{Abdominoperineal resection, left hemicolectomy, low anterior resection, not specified, other, polypectomy, right hemicolectomy, sigmoidectomy, total abdominal colectomy, transanal disk excision, transverse colectomy, or not reported} \\
\hhline{--}
\multicolumn{1}{|p{1.0000in}}{Kidney} & 
\multicolumn{1}{|p{5.3000in}|}{Total nephrectomy, partial nephrectomy, radical nephrectomy, other, or not reported} \\
\hhline{--}
\multicolumn{1}{|p{1.0000in}}{Laterality} & 
\multicolumn{1}{p{5.3000in}|}{} \\
\hhline{--}
\multicolumn{1}{|p{1.0000in}}{Colon} & 
\multicolumn{1}{|p{5.3000in}|}{Not applicable to colon cancer} \\
\hhline{--}
\multicolumn{1}{|p{1.0000in}}{Kidney} & 
\multicolumn{1}{|p{5.3000in}|}{Left, right, or not reported} \\
\hhline{--}
\multicolumn{1}{|p{1.0000in}}{Grade} & 
\multicolumn{1}{p{5.3000in}|}{} \\
\hhline{--}
\multicolumn{1}{|p{1.0000in}}{Kidney, Colon} & 
\multicolumn{1}{|p{5.3000in}|}{Grade 1, 2, 3, 4, not applicable, or not reported} \\
\hhline{--}
\multicolumn{1}{|p{1.0000in}}{Lymphovascular Invasion} & 
\multicolumn{1}{p{5.3000in}|}{} \\
\hhline{--}
\multicolumn{1}{|p{1.0000in}}{Kidney, Colon} & 
\multicolumn{1}{|p{5.3000in}|}{Present, absent, or not reported} \\
\hhline{--}
\multicolumn{1}{|p{1.0000in}}{Perineural Invasion} & 
\multicolumn{1}{p{5.3000in}|}{} \\
\hhline{--}
\multicolumn{1}{|p{1.0000in}}{Colon} & 
\multicolumn{1}{|p{5.3000in}|}{Present, absent, or not reported} \\
\hhline{--}
\multicolumn{1}{|p{1.0000in}}{Kidney } & 
\multicolumn{1}{|p{5.3000in}|}{Not applicable for kidney cancer} \\
\hhline{--}

\end{longtable}


\begin{table*}[t]
\centering

\begin{tabular}{p{0.58in}p{0.58in}p{0.58in}p{0.58in}p{0.58in}p{0.58in}p{0.58in}p{0.58in}p{0.58in}}


\hline
\multicolumn{1}{|p{0.58in}}{\textbf{Colon}} & 
\multicolumn{1}{p{0.58in}}{} & 
\multicolumn{1}{p{0.58in}}{} & 
\multicolumn{1}{p{0.58in}}{} & 
\multicolumn{1}{p{0.58in}}{} & 
\multicolumn{1}{p{0.58in}}{} & 
\multicolumn{1}{p{0.58in}}{} & 
\multicolumn{1}{p{0.58in}}{} & 
\multicolumn{1}{p{0.58in}|}{} \\
\hhline{---------}
\multicolumn{1}{|p{0.58in}}{} & 
\multicolumn{1}{|p{0.58in}}{HAN} & 
\multicolumn{1}{|p{0.58in}}{RF} & 
\multicolumn{1}{|p{0.58in}}{SVM} & 
\multicolumn{1}{|p{0.58in}}{Boost} & 
\multicolumn{1}{|p{0.58in}}{Logistic} & 
\multicolumn{1}{|p{0.58in}}{Rules} & 
\multicolumn{1}{|p{0.58in}}{SLA} & 
\multicolumn{1}{|p{0.58in}|}{Oracle} \\
\hhline{---------}
\multicolumn{1}{|p{0.58in}}{Micro-F1} & 
\multicolumn{1}{p{0.58in}}{} & 
\multicolumn{1}{p{0.58in}}{} & 
\multicolumn{1}{p{0.58in}}{} & 
\multicolumn{1}{p{0.58in}}{} & 
\multicolumn{1}{p{0.58in}}{} & 
\multicolumn{1}{p{0.58in}}{} & 
\multicolumn{1}{p{0.58in}}{} & 
\multicolumn{1}{p{0.58in}|}{} \\
\hhline{---------}
\multicolumn{1}{|p{0.58in}}{32} & 
\multicolumn{1}{|p{0.58in}}{.45} & 
\multicolumn{1}{|p{0.58in}}{.57} & 
\multicolumn{1}{|p{0.58in}}{.51} & 
\multicolumn{1}{|p{0.58in}}{.69} & 
\multicolumn{1}{|p{0.58in}}{.63} & 
\multicolumn{1}{|p{0.58in}}{.75} & 
\multicolumn{1}{|p{0.58in}}{\textbf{.80}} & 
\multicolumn{1}{|p{0.58in}|}{.80} \\
\hhline{---------}
\multicolumn{1}{|p{0.58in}}{64} & 
\multicolumn{1}{|p{0.58in}}{.50} & 
\multicolumn{1}{|p{0.58in}}{.62} & 
\multicolumn{1}{|p{0.58in}}{.57} & 
\multicolumn{1}{|p{0.58in}}{.79} & 
\multicolumn{1}{|p{0.58in}}{.70} & 
\multicolumn{1}{|p{0.58in}}{.79} & 
\multicolumn{1}{|p{0.58in}}{\textbf{.84}} & 
\multicolumn{1}{|p{0.58in}|}{.84} \\
\hhline{---------}
\multicolumn{1}{|p{0.58in}}{128} & 
\multicolumn{1}{|p{0.58in}}{.60} & 
\multicolumn{1}{|p{0.58in}}{.69} & 
\multicolumn{1}{|p{0.58in}}{.62} & 
\multicolumn{1}{|p{0.58in}}{.84} & 
\multicolumn{1}{|p{0.58in}}{.79} & 
\multicolumn{1}{|p{0.58in}}{.83} & 
\multicolumn{1}{|p{0.58in}}{\textbf{.87}} & 
\multicolumn{1}{|p{0.58in}|}{.88} \\
\hhline{---------}
\multicolumn{1}{|p{0.58in}}{186} & 
\multicolumn{1}{|p{0.58in}}{.61} & 
\multicolumn{1}{|p{0.58in}}{.73} & 
\multicolumn{1}{|p{0.58in}}{.68} & 
\multicolumn{1}{|p{0.58in}}{.86} & 
\multicolumn{1}{|p{0.58in}}{.83} & 
\multicolumn{1}{|p{0.58in}}{.84} & 
\multicolumn{1}{|p{0.58in}}{\textbf{.89}} & 
\multicolumn{1}{|p{0.58in}|}{.89} \\
\hhline{---------}
\multicolumn{1}{|p{0.58in}}{Macro-F1} & 
\multicolumn{1}{p{0.58in}}{} & 
\multicolumn{1}{p{0.58in}}{} & 
\multicolumn{1}{p{0.58in}}{} & 
\multicolumn{1}{p{0.58in}}{} & 
\multicolumn{1}{p{0.58in}}{} & 
\multicolumn{1}{p{0.58in}}{} & 
\multicolumn{1}{p{0.58in}}{} & 
\multicolumn{1}{p{0.58in}|}{} \\
\hhline{---------}
\multicolumn{1}{|p{0.58in}}{32} & 
\multicolumn{1}{|p{0.58in}}{.18} & 
\multicolumn{1}{|p{0.58in}}{.22} & 
\multicolumn{1}{|p{0.58in}}{.21} & 
\multicolumn{1}{|p{0.58in}}{.32} & 
\multicolumn{1}{|p{0.58in}}{.28} & 
\multicolumn{1}{|p{0.58in}}{.43} & 
\multicolumn{1}{|p{0.58in}}{\textbf{.50}} & 
\multicolumn{1}{|p{0.58in}|}{.50} \\
\hhline{---------}
\multicolumn{1}{|p{0.58in}}{64} & 
\multicolumn{1}{|p{0.58in}}{.19} & 
\multicolumn{1}{|p{0.58in}}{.29} & 
\multicolumn{1}{|p{0.58in}}{.24} & 
\multicolumn{1}{|p{0.58in}}{.46} & 
\multicolumn{1}{|p{0.58in}}{.37} & 
\multicolumn{1}{|p{0.58in}}{.52} & 
\multicolumn{1}{|p{0.58in}}{\textbf{.60}} & 
\multicolumn{1}{|p{0.58in}|}{.59} \\
\hhline{---------}
\multicolumn{1}{|p{0.58in}}{128} & 
\multicolumn{1}{|p{0.58in}}{.25} & 
\multicolumn{1}{|p{0.58in}}{.34} & 
\multicolumn{1}{|p{0.58in}}{.30} & 
\multicolumn{1}{|p{0.58in}}{.57} & 
\multicolumn{1}{|p{0.58in}}{.53} & 
\multicolumn{1}{|p{0.58in}}{.56} & 
\multicolumn{1}{|p{0.58in}}{\textbf{.66}} & 
\multicolumn{1}{|p{0.58in}|}{.67} \\
\hhline{---------}
\multicolumn{1}{|p{0.58in}}{186} & 
\multicolumn{1}{|p{0.58in}}{.37} & 
\multicolumn{1}{|p{0.58in}}{.40} & 
\multicolumn{1}{|p{0.58in}}{.35} & 
\multicolumn{1}{|p{0.58in}}{.62} & 
\multicolumn{1}{|p{0.58in}}{.59} & 
\multicolumn{1}{|p{0.58in}}{.59} & 
\multicolumn{1}{|p{0.58in}}{\textbf{.69}} & 
\multicolumn{1}{|p{0.58in}|}{.70} \\
\hhline{---------}
\multicolumn{1}{|p{0.58in}}{\textbf{Kidney}} & 
\multicolumn{1}{p{0.58in}}{} & 
\multicolumn{1}{p{0.58in}}{} & 
\multicolumn{1}{p{0.58in}}{} & 
\multicolumn{1}{p{0.58in}}{} & 
\multicolumn{1}{p{0.58in}}{} & 
\multicolumn{1}{p{0.58in}}{} & 
\multicolumn{1}{p{0.58in}}{} & 
\multicolumn{1}{p{0.58in}|}{} \\
\hhline{---------}
\multicolumn{1}{|p{0.58in}}{Micro-F1} & 
\multicolumn{1}{p{0.58in}}{} & 
\multicolumn{1}{p{0.58in}}{} & 
\multicolumn{1}{p{0.58in}}{} & 
\multicolumn{1}{p{0.58in}}{} & 
\multicolumn{1}{p{0.58in}}{} & 
\multicolumn{1}{p{0.58in}}{} & 
\multicolumn{1}{p{0.58in}}{} & 
\multicolumn{1}{p{0.58in}|}{} \\
\hhline{---------}
\multicolumn{1}{|p{0.58in}}{32} & 
\multicolumn{1}{|p{0.58in}}{.54} & 
\multicolumn{1}{|p{0.58in}}{.57} & 
\multicolumn{1}{|p{0.58in}}{.56} & 
\multicolumn{1}{|p{0.58in}}{.73} & 
\multicolumn{1}{|p{0.58in}}{.66} & 
\multicolumn{1}{|p{0.58in}}{.79} & 
\multicolumn{1}{|p{0.58in}}{\textbf{.80}} & 
\multicolumn{1}{|p{0.58in}|}{.85} \\
\hhline{---------}
\multicolumn{1}{|p{0.58in}}{64} & 
\multicolumn{1}{|p{0.58in}}{.54} & 
\multicolumn{1}{|p{0.58in}}{.67} & 
\multicolumn{1}{|p{0.58in}}{.60} & 
\multicolumn{1}{|p{0.58in}}{.80} & 
\multicolumn{1}{|p{0.58in}}{.75} & 
\multicolumn{1}{|p{0.58in}}{.81} & 
\multicolumn{1}{|p{0.58in}}{\textbf{.83}} & 
\multicolumn{1}{|p{0.58in}|}{.86} \\
\hhline{---------}
\multicolumn{1}{|p{0.58in}}{128} & 
\multicolumn{1}{|p{0.58in}}{.63} & 
\multicolumn{1}{|p{0.58in}}{.75} & 
\multicolumn{1}{|p{0.58in}}{.68} & 
\multicolumn{1}{|p{0.58in}}{\textbf{.85}} & 
\multicolumn{1}{|p{0.58in}}{.81} & 
\multicolumn{1}{|p{0.58in}}{.83} & 
\multicolumn{1}{|p{0.58in}}{.84} & 
\multicolumn{1}{|p{0.58in}|}{.87} \\
\hhline{---------}
\multicolumn{1}{|p{0.58in}}{186} & 
\multicolumn{1}{|p{0.58in}}{.71} & 
\multicolumn{1}{|p{0.58in}}{.77} & 
\multicolumn{1}{|p{0.58in}}{.71} & 
\multicolumn{1}{|p{0.58in}}{\textbf{.86}} & 
\multicolumn{1}{|p{0.58in}}{.83} & 
\multicolumn{1}{|p{0.58in}}{.84} & 
\multicolumn{1}{|p{0.58in}}{.85} & 
\multicolumn{1}{|p{0.58in}|}{.88} \\
\hhline{---------}
\multicolumn{1}{|p{0.58in}}{Macro-F1} & 
\multicolumn{1}{p{0.58in}}{} & 
\multicolumn{1}{p{0.58in}}{} & 
\multicolumn{1}{p{0.58in}}{} & 
\multicolumn{1}{p{0.58in}}{} & 
\multicolumn{1}{p{0.58in}}{} & 
\multicolumn{1}{p{0.58in}}{} & 
\multicolumn{1}{p{0.58in}}{} & 
\multicolumn{1}{p{0.58in}|}{} \\
\hhline{---------}
\multicolumn{1}{|p{0.58in}}{32} & 
\multicolumn{1}{|p{0.58in}}{.25} & 
\multicolumn{1}{|p{0.58in}}{.28} & 
\multicolumn{1}{|p{0.58in}}{.29} & 
\multicolumn{1}{|p{0.58in}}{.37} & 
\multicolumn{1}{|p{0.58in}}{.35} & 
\multicolumn{1}{|p{0.58in}}{.46} & 
\multicolumn{1}{|p{0.58in}}{\textbf{.48}} & 
\multicolumn{1}{|p{0.58in}|}{.51} \\
\hhline{---------}
\multicolumn{1}{|p{0.58in}}{64} & 
\multicolumn{1}{|p{0.58in}}{.27} & 
\multicolumn{1}{|p{0.58in}}{.35} & 
\multicolumn{1}{|p{0.58in}}{.30} & 
\multicolumn{1}{|p{0.58in}}{.46} & 
\multicolumn{1}{|p{0.58in}}{.42} & 
\multicolumn{1}{|p{0.58in}}{.49} & 
\multicolumn{1}{|p{0.58in}}{\textbf{.52}} & 
\multicolumn{1}{|p{0.58in}|}{.54} \\
\hhline{---------}
\multicolumn{1}{|p{0.58in}}{128} & 
\multicolumn{1}{|p{0.58in}}{.36} & 
\multicolumn{1}{|p{0.58in}}{.42} & 
\multicolumn{1}{|p{0.58in}}{.38} & 
\multicolumn{1}{|p{0.58in}}{.52} & 
\multicolumn{1}{|p{0.58in}}{.49} & 
\multicolumn{1}{|p{0.58in}}{.51} & 
\multicolumn{1}{|p{0.58in}}{\textbf{.54}} & 
\multicolumn{1}{|p{0.58in}|}{.55} \\
\hhline{---------}
\multicolumn{1}{|p{0.58in}}{186} & 
\multicolumn{1}{|p{0.58in}}{.47} & 
\multicolumn{1}{|p{0.58in}}{.46} & 
\multicolumn{1}{|p{0.58in}}{.42} & 
\multicolumn{1}{|p{0.58in}}{.54} & 
\multicolumn{1}{|p{0.58in}}{.51} & 
\multicolumn{1}{|p{0.58in}}{.54} & 
\multicolumn{1}{|p{0.58in}}{\textbf{.56}} & 
\multicolumn{1}{|p{0.58in}|}{.56} \\
\hhline{---------}


\end{tabular}

\caption{Average micro-f1 and macro-f1 performance across attributes of different methods as a function of 32, 64, 128, 186 labeled examples on colon and kidney cancer. Highest performing non-oracle method is bolded. }
\label{table:results_table}

\end{table*}

\subsection{Dataset Statistics}
Dataset Statistics:
In this section we discuss various statistics of our dataset. The average number of tokens in our colon cancer reports is 1060 tokens. The lower 5\% quantile of lengths is 457 tokens and the upper 95\% quantile of lengths is 1826 tokens. The average number of tokens in our kidney cancer reports is 806 tokens and the lower 5\% quantile of lengths is 393 and the upper 95\% quantile is 1485. 

\subsection{Annotator reliability check}
To check the annotation quality, a second trained individual verified each original annotation across the train, validation, and test sets. We present two annotator agreement metrics on a per field and cancer basis for the compiled reports: the fraction of agreements across annotations and the cohen kappa metric (Table 5). For a majority of the fields for both colon and kidney, the agreement fraction and cohen kappa metric are well above 0.9 suggesting a high level of annotator agreement. Tumor site for kidney cancer is the only exception with agreement metrics at 0.88 and 0.856. 

\subsection{Data Preprocessing}

For all methods, we replace all words that occur fewer than two times in the training data with a special $<$UNK$>$ token, and remove commas, backslashes, semi-colons, tildes, periods, and the word $``$null$"$  from each report in the corpus. For colons, forward slashes, parentheses, plus, and equal signs, we added a space before and after the character. The spaces were artificially added to preserve semantic value important to the task. For instance, colons often appear in the synoptic comment, and so if an n-gram contains a colon, it can indicate that the n-gram contains important information. 

\subsection{Ablation Experiments}

For our ablation experiments, we investigate the relative contribution of each component in our model. 
\subsubsection*{No weighting:} Here we investigate if weighting the features in each line by the classifier scores increases performance compared to weighting the features in each line by one.
\subsubsection*{No joining:} Here we investigate how joining affects the results when information spans multiple lines. Instead of conjoining lines that occur adjacent to each other, we leave them as separate lines for our final classifier. 

\subsubsection*{No weighting and no joining:}

Here we neither weight the features vectors representing each line nor do we join adjacent predicted lines.

\subsection{Ablation Results}

\begin{figure*}[ht]
	\centering
		\includegraphics[width=1\textwidth]{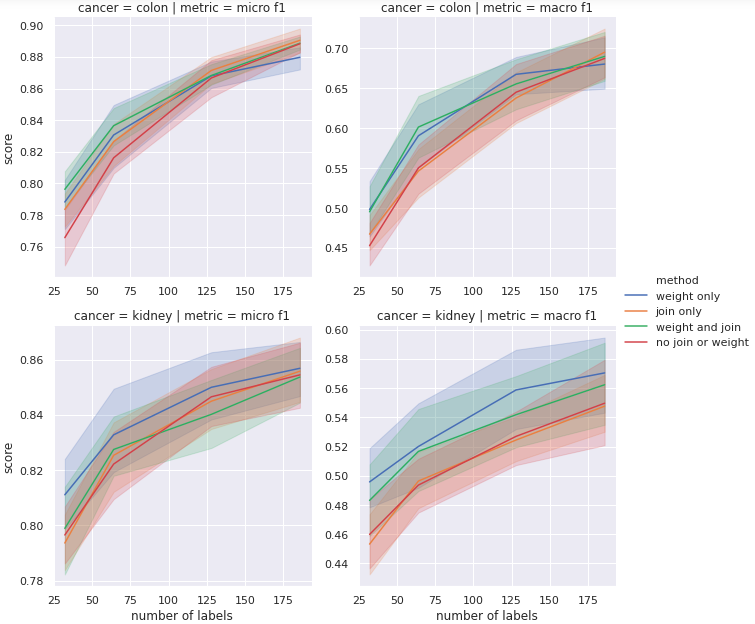}
	\caption[width=.5\textwidth]{Ablation studies for SLA measuring the average micro-f1 and macro-f1 performance across attributes of different methods as a function of 32, 64, 128, 186 labeled examples on colon cancer and kidney cancer pathology reports. We investigate the impact of joining adjacent selected lines prior to featurization as well as the impact of weighting the features by the line classifier scores. We present the mean result across 10 random shufflings of the data with 95$\%$  bootstrap confidence intervals.}
\label{fig:ablations}

\end{figure*}

We plot the results of our ablation experiments in Figure~\ref{fig:ablations}, using the same setup as our main result where we have training set sizes of 32, 64, 128, and 186 with 4-fold cross validation. Again, we take the average of 10 runs where we reshuffle the data and generate new splits each time and compute 95$\%$  confidence intervals for all methods using bootstrap resampling. We see mixed results for joining adjacent predicted lines; it appears to be inconsequential for colon cancer and detrimental for kidney cancer.  However, weighting the features by line predictor seems beneficial for the macro-f1 scores. This seems to suggest that weighting helps primarily for rare classes since the macro-f1 score weights the f1 scores of each class equally.

\subsection{Cancer Stage}
{
We do not consider pathologic stage as a classification task in this paper, as each stage is encoded in a single token, such as pT2aN0M0. The letters T, N, and M denote the pathologic stage type, while 2a and 0 that follow immediately denote the class for each stage type. Classifier methods using full words as tokens are not well suited for this task, because the number of possible token encodings is large (equal to the number of classes for the T-stage multiplied by the number of classes for the N and M-stages). It may even be the case that a certain encoding may show up in the test set and does not show up in the train set if a certain combination of the stages is new. Though the details and results are outside the scope of this paper, we handle cancer staging by first extracting the stage token (eg pT2aN0M0) and then determining the values for T, N, and M stages using a regular expression rule-based method. The results on the full training data are presented in Table A.6.}

\begin{table*}
\centering
\begin{tabular}{|p{1.1in}|p{0.58in}|p{0.5in}|p{0.58in}|p{0.97in}|p{0.55in}|p{0.55in}|p{0.38in}|}
\hline
\multicolumn{1}{p{1.1in}}{Attribute} & 
\multicolumn{1}{p{0.58in}}{Histologic Grade} & 
\multicolumn{1}{p{0.5in}}{Histologic Type} & 
\multicolumn{1}{p{0.58in}}{Perineural invasion} & 
\multicolumn{1}{p{0.97in}}{Lymphovascular invasion} & 
\multicolumn{1}{p{0.55in}}{Procedure} & 
\multicolumn{1}{p{0.55in}}{Tumor Site} & 
\multicolumn{1}{p{0.38in}}{Total} \\
\hline
\multicolumn{1}{p{1.1in}}{Attribute Qualification Error} & 
\multicolumn{1}{p{0.58in}}{1} & 
\multicolumn{1}{p{0.5in}}{0} & 
\multicolumn{1}{p{0.58in}}{0} & 
\multicolumn{1}{p{0.97in}}{0} & 
\multicolumn{1}{p{0.55in}}{0} & 
\multicolumn{1}{p{0.55in}}{0} & 
\multicolumn{1}{p{0.38in}}{1} \\
\hline
\multicolumn{1}{p{1.1in}}{Rare \par phrasing} & 
\multicolumn{1}{p{0.58in}}{0} & 
\multicolumn{1}{p{0.5in}}{0} & 
\multicolumn{1}{p{0.58in}}{0} & 
\multicolumn{1}{p{0.97in}}{\textbf{1}} & 
\multicolumn{1}{p{0.55in}}{3} & 
\multicolumn{1}{p{0.55in}}{0} & 
\multicolumn{1}{p{0.38in}}{4} \\
\hline
\multicolumn{1}{p{1.1in}}{Irrelevant Lines Error} & 
\multicolumn{1}{p{0.58in}}{1} & 
\multicolumn{1}{p{0.5in}}{0} & 
\multicolumn{1}{p{0.58in}}{0} & 
\multicolumn{1}{p{0.97in}}{0} & 
\multicolumn{1}{p{0.55in}}{5} & 
\multicolumn{1}{p{0.55in}}{0} & 
\multicolumn{1}{p{0.38in}}{6} \\
\hline
\multicolumn{1}{p{1.1in}}{Annotator Error Error} & 
\multicolumn{1}{p{0.58in}}{3} & 
\multicolumn{1}{p{0.5in}}{\textbf{1}} & 
\multicolumn{1}{p{0.58in}}{\textbf{1}} & 
\multicolumn{1}{p{0.97in}}{0} & 
\multicolumn{1}{p{0.55in}}{5} & 
\multicolumn{1}{p{0.55in}}{0} & 
\multicolumn{1}{p{0.38in}}{10} \\
\hline
\multicolumn{1}{p{1.1in}}{Multi-label Error} & 
\multicolumn{1}{p{0.58in}}{\textbf{6}} & 
\multicolumn{1}{p{0.5in}}{0} & 
\multicolumn{1}{p{0.58in}}{0} & 
\multicolumn{1}{p{0.97in}}{0} & 
\multicolumn{1}{p{0.55in}}{0} & 
\multicolumn{1}{p{0.55in}}{\textbf{6}} & 
\multicolumn{1}{p{0.38in}}{\textbf{12}} \\
\hline
\multicolumn{1}{p{1.1in}}{Unknown error} & 
\multicolumn{1}{p{0.58in}}{1} & 
\multicolumn{1}{p{0.5in}}{0} & 
\multicolumn{1}{p{0.58in}}{0} & 
\multicolumn{1}{p{0.97in}}{0} & 
\multicolumn{1}{p{0.55in}}{\textbf{6}} & 
\multicolumn{1}{p{0.55in}}{0} & 
\multicolumn{1}{p{0.38in}}{7} \\

\hline
\multicolumn{1}{p{1.1in}}{Total by attribute} & 
\multicolumn{1}{p{0.58in}}{12} & 
\multicolumn{1}{p{0.5in}}{1} & 
\multicolumn{1}{p{0.58in}}{1} & 
\multicolumn{1}{p{0.97in}}{1} & 
\multicolumn{1}{p{0.55in}}{\textbf{19}} & 
\multicolumn{1}{p{0.55in}}{6} & 
\multicolumn{1}{p{0.38in}}{40} \\
\hline
\end{tabular}
\caption{Error analysis: Colon cancer}
\label{table:err_colon}

\end{table*}

\begin{table*}
\centering

\begin{tabular}{p{1.16in}p{0.63in}p{0.54in}p{0.57in}p{0.96in}p{0.55in}p{0.34in}p{0.31in}}
\hline
\multicolumn{1}{p{1.16in}}{Attribute} & 
\multicolumn{1}{p{0.63in}}{Histologic Grade} & 
\multicolumn{1}{p{0.54in}}{Histologic Type} & 
\multicolumn{1}{p{0.57in}}{Specimen Laterality} & 
\multicolumn{1}{p{0.96in}}{Lymphovascular invasion} & 
\multicolumn{1}{p{0.55in}}{Procedure} & 
\multicolumn{1}{p{0.34in}}{Tumor Site} & 
\multicolumn{1}{p{0.31in}}{Total} \\
\hline
\multicolumn{1}{p{1.16in}}{Attribute Qualification Error} & 
\multicolumn{1}{p{0.63in}}{0} & 
\multicolumn{1}{p{0.54in}}{0} & 
\multicolumn{1}{p{0.57in}}{0} & 
\multicolumn{1}{p{0.96in}}{0} & 
\multicolumn{1}{p{0.55in}}{0} & 
\multicolumn{1}{p{0.34in}}{0} & 
\multicolumn{1}{p{0.31in}}{0} \\
\hline
\multicolumn{1}{p{1.16in}}{Rare \par phrasing} & 
\multicolumn{1}{p{0.63in}}{0} & 
\multicolumn{1}{p{0.54in}}{0} & 
\multicolumn{1}{p{0.57in}}{0} & 
\multicolumn{1}{p{0.96in}}{\textbf{1}} & 
\multicolumn{1}{p{0.55in}}{\textbf{5}} & 
\multicolumn{1}{p{0.34in}}{0} & 
\multicolumn{1}{p{0.31in}}{6} \\
\hline
\multicolumn{1}{p{1.16in}}{Irrelevant Lines Error} & 
\multicolumn{1}{p{0.63in}}{\textbf{1}} & 
\multicolumn{1}{p{0.54in}}{0} & 
\multicolumn{1}{p{0.57in}}{0} & 
\multicolumn{1}{p{0.96in}}{\textbf{1}} & 
\multicolumn{1}{p{0.55in}}{1} & 
\multicolumn{1}{p{0.34in}}{1} & 
\multicolumn{1}{p{0.31in}}{4} \\
\hline
\multicolumn{1}{p{1.16in}}{Annotator Error} & 
\multicolumn{1}{p{0.63in}}{\textbf{1}} & 
\multicolumn{1}{p{0.54in}}{2} & 
\multicolumn{1}{p{0.57in}}{0} & 
\multicolumn{1}{p{0.96in}}{\textbf{1}} & 
\multicolumn{1}{p{0.55in}}{1} & 
\multicolumn{1}{p{0.34in}}{0} & 
\multicolumn{1}{p{0.31in}}{5} \\
\hline
\multicolumn{1}{p{1.16in}}{Multi-label Error} & 
\multicolumn{1}{p{0.63in}}{0} & 
\multicolumn{1}{p{0.54in}}{\textbf{4}} & 
\multicolumn{1}{p{0.57in}}{0} & 
\multicolumn{1}{p{0.96in}}{0} & 
\multicolumn{1}{p{0.55in}}{2} & 
\multicolumn{1}{p{0.34in}}{\textbf{6}} & 
\multicolumn{1}{p{0.31in}}{\textbf{12}} \\
\hline
\multicolumn{1}{p{1.16in}}{Unknown error} & 
\multicolumn{1}{p{0.63in}}{\textbf{1}} & 
\multicolumn{1}{p{0.54in}}{\textbf{4}} & 
\multicolumn{1}{p{0.57in}}{0} & 
\multicolumn{1}{p{0.96in}}{\textbf{1}} & 
\multicolumn{1}{p{0.55in}}{1} & 
\multicolumn{1}{p{0.34in}}{5} & 
\multicolumn{1}{p{0.31in}}{\textbf{12}} \\
\hline
\multicolumn{1}{p{1.16in}}{Total by attribute } & 
\multicolumn{1}{p{0.63in}}{3} & 
\multicolumn{1}{p{0.54in}}{10} & 
\multicolumn{1}{p{0.57in}}{0} & 
\multicolumn{1}{p{0.96in}}{4} & 
\multicolumn{1}{p{0.55in}}{10} & 
\multicolumn{1}{p{0.34in}}{\textbf{12}} & 
\multicolumn{1}{p{0.31in}}{39} \\
\hline

\end{tabular}

\caption{Error analysis: Kidney cancer}
\label{table:err_kidney}
\end{table*}

\begin{table*}[t]
\centering

\begin{tabular}{p{0.8in}p{0.8in}p{0.8in}}

\hline
\multicolumn{1}{|p{1.25in}}{\textbf{Colon}} & 
\multicolumn{1}{p{0.8in}}{} & 
\multicolumn{1}{p{0.8in}|}{} \\
\hhline{---}
\multicolumn{1}{|p{1.25in}}{Field} & 
\multicolumn{1}{|p{0.8in}}{Agreement fraction} & 
\multicolumn{1}{|p{0.8in}|}{Cohen kappa} \\
\hhline{---}
\multicolumn{1}{|p{1.25in}}{Histologic grade} & 
\multicolumn{1}{|p{0.8in}}{.976} & 
\multicolumn{1}{|p{0.8in}|}{.966} \\
\hhline{---}
\multicolumn{1}{|p{1.25in}}{Histologic type} & 
\multicolumn{1}{|p{0.80in}}{.996} & 
\multicolumn{1}{|p{0.80in}|}{.979} \\
\hhline{---}
\multicolumn{1}{|p{1.25in}}{Perineural invasion} & 
\multicolumn{1}{|p{0.80in}}{.980} & 
\multicolumn{1}{|p{0.80in}|}{.937} \\
\hhline{---}
\multicolumn{1}{|p{1.25in}}{Lymphovascular invasion} & 
\multicolumn{1}{|p{0.80in}}{.988} & 
\multicolumn{1}{|p{0.80in}|}{.973} \\
\hhline{---}
\multicolumn{1}{|p{1.25in}}{Procedure} & 
\multicolumn{1}{|p{0.80in}}{.960} & 
\multicolumn{1}{|p{0.80in}|}{.951} \\
\hhline{---}
\multicolumn{1}{|p{1.25in}}{Tumor site} & 
\multicolumn{1}{|p{0.80in}}{.948} & 
\multicolumn{1}{|p{0.80in}|}{.938} \\
\hhline{---}
\multicolumn{1}{|p{1.25in}}{\textbf{Kidney}} & 
\multicolumn{1}{p{0.8in}}{} & 
\multicolumn{1}{p{0.8in}|}{} \\
\hhline{---}
\multicolumn{1}{|p{1.25in}}{Field} & 
\multicolumn{1}{|p{0.8in}}{Agreement fraction} & 
\multicolumn{1}{|p{0.8in}|}{Cohen kappa} \\
\hhline{---}
\multicolumn{1}{|p{1.25in}}{Histologic grade} & 
\multicolumn{1}{|p{0.8in}}{.948} & 
\multicolumn{1}{|p{0.8in}|}{.930} \\
\hhline{---}
\multicolumn{1}{|p{1.25in}}{Histologic type} & 
\multicolumn{1}{|p{0.80in}}{.964} & 
\multicolumn{1}{|p{0.80in}|}{.947} \\
\hhline{---}
\multicolumn{1}{|p{1.25in}}{Lymphovascular invasion} & 
\multicolumn{1}{|p{0.80in}}{.984} & 
\multicolumn{1}{|p{0.80in}|}{.962} \\
\hhline{---}
\multicolumn{1}{|p{1.25in}}{Procedure} & 
\multicolumn{1}{|p{0.80in}}{.960} & 
\multicolumn{1}{|p{0.80in}|}{.931} \\
\hhline{---}
\multicolumn{1}{|p{1.25in}}{Specimen laterality} & 
\multicolumn{1}{|p{0.80in}}{.984} & 
\multicolumn{1}{|p{0.80in}|}{.970} \\
\hhline{---}
\multicolumn{1}{|p{1.25in}}{Tumor site} & 
\multicolumn{1}{|p{0.80in}}{.880} & 
\multicolumn{1}{|p{0.80in}|}{.856} \\
\hhline{---}

\end{tabular}

\caption{Annotator agreement metrics across colon and kidney cancers }
\label{table:results_table}

\end{table*}

\begin{table*}[t]
\centering

\begin{tabular}{p{0.8in}p{0.8in}p{0.8in}}

\hline
\multicolumn{1}{|p{1.25in}}{\textbf{Colon}} & 
\multicolumn{1}{p{0.8in}}{} & 
\multicolumn{1}{p{0.8in}|}{} \\
\hhline{---}
\multicolumn{1}{|p{1.25in}}{Field} & 
\multicolumn{1}{|p{0.8in}}{Micro-f1} & 
\multicolumn{1}{|p{0.8in}|}{Macro-F1} \\
\hhline{---}
\multicolumn{1}{|p{1.25in}}{T stage} & 
\multicolumn{1}{|p{0.8in}}{.968} & 
\multicolumn{1}{|p{0.8in}|}{.741} \\
\hhline{---}
\multicolumn{1}{|p{1.25in}}{N stage} & 
\multicolumn{1}{|p{0.80in}}{.968} & 
\multicolumn{1}{|p{0.80in}|}{.811} \\
\hhline{---}
\multicolumn{1}{|p{1.25in}}{M stage} & 
\multicolumn{1}{|p{0.80in}}{.921} & 
\multicolumn{1}{|p{0.80in}|}{.529} \\
\hhline{---}
\multicolumn{1}{|p{1.25in}}{\textbf{Kidney}} & 
\multicolumn{1}{p{0.8in}}{} & 
\multicolumn{1}{p{0.8in}|}{} \\
\hhline{---}
\multicolumn{1}{|p{1.25in}}{Field} & 
\multicolumn{1}{|p{0.8in}}{Micro-f1} & 
\multicolumn{1}{|p{0.8in}|}{Macro-F1} \\
\hhline{---}
\multicolumn{1}{|p{1.25in}}{T stage} & 
\multicolumn{1}{|p{0.8in}}{.936} & 
\multicolumn{1}{|p{0.8in}|}{.680} \\
\hhline{---}
\multicolumn{1}{|p{1.25in}}{N stage} & 
\multicolumn{1}{|p{0.80in}}{.968} & 
\multicolumn{1}{|p{0.80in}|}{.663} \\
\hhline{---}
\multicolumn{1}{|p{1.25in}}{M stage} & 
\multicolumn{1}{|p{0.80in}}{.952} & 
\multicolumn{1}{|p{0.80in}|}{.438} \\
\hhline{---}
\end{tabular}

\caption{Micro-F1 and macro-F1 scores for pathology T, N, and M stages using the full training data}
\label{table:results_table}

\end{table*}

\end{document}